\documentclass{article}

\usepackage{PRIMEarxiv}

\usepackage[utf8]{inputenc} 
\usepackage[T1]{fontenc}    
\usepackage{hyperref}       
\usepackage{url}            
\usepackage{booktabs}       
\usepackage{amsfonts}       
\usepackage{nicefrac}       
\usepackage{microtype}      
\usepackage{mhchem}         
\usepackage{lipsum}
\usepackage{fancyhdr}       
\usepackage{graphicx}       
\usepackage{makecell}
\usepackage{tabularx}
\usepackage{booktabs}
\usepackage{makecell}
\usepackage{threeparttable}
\usepackage{graphicx}
\usepackage{subcaption}
\usepackage{setspace}

\graphicspath{{media/}}     

\title{SAMamba3D: Adapting Segment Anything for Generalizable 3D Segmentation of Multiphase Pore-Scale Images}

\author{
  Rui Zhang$^{1,2}$, Xianzhi Song$^{3*}$, Linqi Zhu$^{1*}$, Branko Bijeljic$^{1}$, Gensheng Li$^{3}$ and Martin J. Blunt$^{1}$ \\[0.5em]
    $^{1}$Department of Earth Science and Engineering, Imperial College London, London SW7 2AZ, United Kingdom\\
    $^{2}$College of Artificial Intelligence, China University of Petroleum (Beijing), Beijing 102249, China\\
    $^{3}$College of Petroleum Engineering, China University of Petroleum (Beijing), Beijing 102249, China \\[0.5em]
  \texttt{linqi.zhu@imperial.ac.uk, songxz@cup.edu.cn} 
}

\begin{document}
\maketitle

\begin{abstract}
Reliable segmentation of multiphase pore-scale X-ray images of rocks is necessary to quantify fluid saturation, connectivity, and interfacial geometry. However, current 3D segmentation methods are typically dataset-specific, requiring retraining or extensive fine-tuning whenever rock type, fluid pattern, scanner, or acquisition conditions change. Foundational models such as the Segment Anything Model, SAM, provide strong 2D boundary priors, but they are not directly applicable to 3D data. 
\vspace{6pt}

We present SAMamba3D, \url{https://github.com/ImperialCollegeLondon/SAMamba-3D},  a parameter-efficient framework that adapts a largely frozen SAM encoder to generalizable 3D pore-scale segmentation by coupling it with Mamba-based volumetric context modeling and progressive cross-scale feature interaction. For sandstone and carbonate datasets, with different fluids, wettability, and scanning conditions, SAMamba3D matches or outperforms current 3D baselines while reducing the need for case-specific retraining. The resulting segmented images preserve physically meaningful descriptors, including fluid saturation, connectivity, and interface morphology, enabling more reliable and rapid analysis of large 3D multiphase images.
\end{abstract}


\section{Introduction}

Multiphase flow in porous media underlies a wide range of subsurface processes of major scientific and practical importance, including geological CO$_2$ sequestration, underground hydrogen storage, and groundwater remediation \cite{bashir2024co2storage,miocic2023uhsreview}. X-ray micro-computed tomography (micro-CT) has become a key pore-scale imaging tool for investigating these processes because it provides non-destructive three-dimensional, and increasingly time-resolved, observations of pore geometry and fluid configurations at micrometer resolution \cite{wildenschild2013xray,blunt2013porescale}. However, reconstructed micro-CT images are not phase-labeled representations of the porous medium and the fluids in the pore space. They are grayscale attenuation fields that must first be segmented before physically meaningful quantities such as porosity, phase saturation, fluid connectivity, interfacial area, contact angle, and curvature can be estimated \cite{iassonov2009segmentation,schluter2014review,scanziani2017contactangle,huang2021topology,andrew2014contact,alratrout2017automatic,lin2018curvature}. In this setting, segmentation is not merely a preprocessing step: it is the operation that converts raw attenuation contrast into a geometric and topological description suitable for pore-scale analysis. Errors at the voxel scale can alter connected pathways, and distort interfacial geometry, causing errors that propagate directly into pore-network extraction, numerical flow simulation, and physical interpretation.

Accurate segmentation of multiphase pore-scale images remains difficult because grayscale distributions often overlap across phases, especially in the presence of imaging noise, reconstruction artifacts, limited spatial resolution, and partial-volume effects near thin wetting layers, narrow pore throats, and complex fluid--fluid or fluid--solid interfaces. Manual and semi-automated workflows, including thresholding and watershed-based methods, remain widely used because they are simple and interpretable, but they are sensitive to user choices and difficult to reproduce consistently across datasets \cite{iassonov2009segmentation,schluter2014review}. CNN-based encoder--decoder and transformer-inspired models have improved automation and local boundary delineation \cite{niu2020digitalrock,alqahtani2022superresolved,wang2024comparativeunet,siavashi2024deepautoencoder,gao2024gradientseg,mahdaviara2023gdl}. However, most existing methods remain strongly tied to the training distribution. Performance often degrades when rock type, wettability, fluid configuration, scanner hardware, or acquisition conditions change. For multiphase pore-scale micro-CT, where cross-dataset heterogeneity is the norm rather than the exception, robust generalization beyond a single dataset remains a central challenge.

Recent progress in vision foundational models suggests a promising alternative. The Segment Anything Model (SAM) \cite{kirillov2023sam} showed that large-scale pre-training can provide transferable segmentation priors, and subsequent studies in medical imaging have demonstrated that such models can often be efficiently adapted to specialized domains through lightweight tuning rather than full retraining \cite{ma2024medsam,wu2023medsamadapter,chen2024masam,yan2024aftersam}. This is appealing for pore-scale micro-CT, where annotated 3D data are expensive to obtain and constructing a separate segmentation pipeline for each rock--fluid system is impractical. However, SAM on its own is not directly applicable to multiphase pore-scale micro-CT images. SAM was designed for 2D color images, whereas pore-scale images are 3D, monochromatic (grayscale only), frequently low-contrast, and strongly affected by reconstruction artifacts and partial-volume effects. Slice-wise transfer can break volumetric continuity and fail to preserve connected pore structures, thin wetting layers, and interfacial topology \cite{wu2023medsamadapter,chen2024masam,yan2024aftersam}. Moreover, ambiguous voxels often cannot be resolved from local boundary appearance alone. They must be interpreted in the context of surrounding pore geometry and the broader spatial organization of fluid phases. Effective adaptation therefore requires 3D consistency, richer structural context, and practical computational efficiency. 
Tang et al. \cite{Tang2025nc} have introduced a deep learning model with domain transfer that has shown promising results for a range of 3D images, including rocks, batteries and fuel cells. However, additional training is still required to segment a new dataset.  Our aim is to provide a method that does not require additional training, albeit for a more restricted range of multiphase images of porous rocks.

In this work, we recast multiphase pore-scale micro-CT segmentation as a foundational-model adaptation problem. We present \textbf{SAMamba3D}, a reusable 3D segmentation framework that extends a largely frozen SAM encoder to volumetric data through lightweight 3D adaptation modules and couples it with a Mamba branch for global structural modeling. Rather than using SAM as a fixed 2D feature extractor followed by simple late fusion, the framework is designed so that SAM-based boundary priors and Mamba-based multi-scale volumetric features interact throughout the segmentation pipeline. The goal is not only to improve segmentation on individual datasets, but also to examine whether foundation-model adaptation can provide a more transferable strategy for multiphase pore-scale image analysis.

Specifically, this work makes three contributions. First, we introduce a parameter-efficient approach for adapting SAM to 3D multiphase pore-scale micro-CT segmentation. Second, we develop a volumetric context-fusion architecture that combines local boundary-aware representations with broader structural reasoning. Third, we evaluate the method not only in terms of segmentation accuracy, but also test cross-dataset generalization and preservation of downstream physical descriptors relevant to pore-scale analysis. These results present that reusable foundational-model adaptation is a viable route toward scalable and physically reliable segmentation of multiphase pore-scale X-ray images.

\section{Methodology}
This section outlines the dataset construction and the training, validation and testing protocol, followed by the preprocessing pipeline, the SAMamba3D architecture, the boundary-aware composite loss, and the training and inference settings. Note that all validation was performed on the training datasets: all tests were performed on datasets that were not used for training and validation.
Given that the central claim of this work is generalization beyond case-specific retraining, the data partitioning and evaluation protocol are presented prior to the model description.

\subsection{Datasets}
\label{sec:2.1sec}
\subsubsection{Dataset Description}
To evaluate segmentation beyond a single rock type or imaging condition, we assembled a curated multi-source dataset of multiphase pore-scale X-ray micro-CT volumes from published studies. The collection covers sandstone, carbonate, and glass bead pack samples, multiple fluid systems, and a range of wettability states acquired from different scanners with varying resolution. Table~\ref{tab:table1} lists the datasets used in this study, including rock type, fluid system, wettability, voxel size, shape of the imaged volumes, and the role of each dataset for either training or testing.

\paragraph{Sandstone and bead-pack datasets}
The sandstone subset includes Bentheimer volumes from oil--brine displacement experiments under water-wet and mixed-wet conditions and at different water fractional flows. There are also mages of Mt. Simon sandstone under supercritical \ce{CO2} (sc\ce{CO2})--brine conditions, and Clashach sandstone imaged during \ce{H2}--brine displacement. We also include a borosilicate bead pack acquired for sc\ce{CO2}--brine flow experiments. These datasets cover relatively homogeneous media, a layered sandstone, and a granular reference material, sampling different pore morphologies and contrast regimes.

\paragraph{Carbonate datasets}
The carbonate subset includes Ketton limestone, Estaillades carbonate, Indiana limestone, and Middle Eastern carbonate samples under weakly water-wet, mixed-wet, and oil-wet conditions. Compared with sandstones, these datasets exhibit substantially stronger structural heterogeneity, a broader range of pore sizes, and more complex interfacial configurations, making them a critical test of transfer beyond narrowly matched training data.

\paragraph{Scope of variation}
Taken together, the dataset spans the principal sources of variation encountered in multiphase pore-scale imaging: rock type, fluid pair, wettability state, scanner-dependent gray-level statistics, and image resolution. This diversity is necessary for evaluating whether a single segmentation framework can be reused across conditions without retraining a separate model for each dataset.

\begin{table*}[htbp]
\centering
\small
\begin{threeparttable}
\caption{Summary of the datasets used in this study. Under the column Role, Train/ Val indicates datasets used for training and validation, while Test represents datasets used to test the segmentation.}
\label{tab:table1}
\setlength{\tabcolsep}{5pt}
\renewcommand{\arraystretch}{1.0}
\begin{tabularx}{\textwidth}{l l l c c c c}
\toprule
\textbf{Dataset} & \textbf{Rock type} & \textbf{Fluid system} & \textbf{Wettability} & \textbf{Voxel size}   ($\mu$m) & \textbf{Volumes} & \textbf{Role} \\ 
\midrule
Bentheimer \cite{lin2018curvature,shojaei2023,lin2019bentheimer}
& Sandstone & Oil--brine
& \makecell{Water-wet\\Water-wet\\ Mixed-wet}
& \makecell{3.56\\ 3.58\\ 3.85}
& \makecell{$1000 \times 1000 \times 3000$\\ $1000 \times 1000 \times 3000$\\ $1000 \times 1000 \times 3000$}
& \makecell{Train/ Val \\ Test \\ Test} \\
\cmidrule(lr){4-7} 
\addlinespace[2pt]
Mt.\ Simon \cite{dalton2018methods}
& Sandstone & scCO$_2$--brine & Water-wet
& \makecell{1.66\\ 2.32}
& \makecell{$400 \times 400 \times 400$\\ $286 \times 286 \times 286$}
& Test \\
\cmidrule(lr){4-7} 
\addlinespace[2pt]
Clashach \cite{jangda2024subsurface}
& Sandstone & H$_2$--brine & Water-wet
& 7 & $800 \times 800 \times 1207$ & Test \\
\cmidrule(lr){4-7} 
\addlinespace[2pt]
Glass Bead Pack \cite{tawfik2022denoising}
& Borosilicate & scCO$_2$--brine & Water-wet
& 15 & $800 \times 800 \times 5$ & Test \\
\cmidrule(lr){4-7} 
\addlinespace[2pt]
Ketton \cite{scanziani2018threephase,singh2018timresolved}
& Carbonate & Oil--brine & Water-wet
& \makecell{5\\ 2\\ 2}
& \makecell{$540 \times 540 \times 2961$\\ $510 \times 410 \times 410$\\ $225 \times 255 \times 365$}
& \makecell{Train/ Val\\ Train/ Val\\ Test} \\
\cmidrule(lr){4-7} 
\addlinespace[2pt]
Estaillades \cite{bultreys2016estaillades,Hussain2025esta}
& Carbonate & Oil--brine & Oil-wet
& \makecell{3.1\\ 2.8}
& \makecell{$1000 \times 1000 \times 1000$\\ $1000 \times 1000 \times 3755$}
& Test \\
\cmidrule(lr){4-7} 
\addlinespace[2pt]
Indiana \cite{alqahtani2022superresolved}
& Carbonate & Oil--brine & Water-wet
& 2.68 & $1520 \times 1520 \times 888$ & Test \\
\cmidrule(lr){4-7} 
\addlinespace[2pt]
Middle Eastern \cite{alhammadi2017mixedwet,alqahtani2022superresolved}
& Carbonate & Oil--brine
& \makecell{Water-wet\\ Mixed-wet\\Water-wet}
& \makecell{2\\2\\ 2.68}
& \makecell{$660 \times 660 \times 601$\\$660 \times 660 \times 601$\\ $1520 \times 1520 \times 1025$}
& \makecell{Train/ Val\\ Test \\ Test} \\
\bottomrule
\end{tabularx}
\end{threeparttable}
\end{table*}

\subsubsection{Data Preprocessing}
Raw micro-CT reconstructions from different scanners and acquisition settings exhibit substantial variation in noise characteristics, gray-level distributions, and effective contrast. To reduce scanner-specific intensity shifts while avoiding information leakage, we apply denoising, percentile-based intensity alignment, and global standardization in that order.

\paragraph{Denoising}
Each reconstructed 3D volume was filtered using a non-local means (NLM) algorithm, which suppresses photon-counting noise and ring artifacts while preserving phase-boundary sharpness \cite{buades2005nlm}. The NLM filter exploits self-similarity across non-adjacent voxel neighborhoods, making it effective for micro-CT data where edges carry critical physical information.

\paragraph{Percentile-based intensity alignment}
Directly combining volumes from different scanners can introduce artificial domain shifts because the absolute gray-level ranges are not comparable. We align each source volume \(I_s\) to a fixed reference intensity range defined from the training data only. Let \(q_s^{(p)}\) and \(q_r^{(p)}\) denote the \(p\)-th intensity percentile of the source and reference volumes, respectively. Using \(p_{\text{low}}=1\) and \(p_{\text{high}}=99\), we apply
\begin{equation}
I_s^{\prime} =
\frac{I_s - q_s^{(p_{\text{low}})}}{q_s^{(p_{\text{high}})} - q_s^{(p_{\text{low}})}}
\cdot
\left(q_r^{(p_{\text{high}})} - q_r^{(p_{\text{low}})}\right)
+
q_r^{(p_{\text{low}})}.
\end{equation}
The reference range is defined as pooled training percentiles and is reused unchanged for validation and test data.

\paragraph{Global standardization}
After percentile alignment, all volumes are standardized using the global mean \(\mu_g\) and standard deviation \(\sigma_g\) computed from the training split only:
\begin{equation}
\hat{I}(v) = \frac{I(v) - \mu_g}{\sigma_g + \epsilon},
\end{equation}
where \(\epsilon = 10^{-8}\) is a numerical stability constant. The same \(\mu_g\) and \(\sigma_g\) are then applied to the validation and test sets.

\subsubsection{Data Augmentation}
To improve robustness under limited labeled data, we apply on-the-fly stochastic augmentation to each sampled 3D patch during training. The augmentation set includes axis-wise random flipping (probability \(0.5\) per axis), random \(90^\circ\) rotations in the axial plane, multiplicative brightness shift (uniform scale factor in \([0.9, 1.1]\)), and additive Gaussian noise with \(\sigma = 0.02\) (probability \(0.3\)). Geometric transformations were applied identically to the image patch and its corresponding label mask, and nearest-neighbor interpolation is used for mask resampling. No augmentation is applied during validation or testing.

\subsubsection{Data Partitioning}
Training is performed on 3D patches of size \(P_d \times P_h \times P_w\), set to \(96 \times 96 \times 96\) in this study. Patches are extracted using a sliding window with stride \(48\) voxels. Patches containing less than \(10\%\) foreground voxels are discarded.
All train, validation and test partitions are defined at the volume-level before patch extraction. No overlapping or adjacent patches from the same physical region are placed in different splits. This avoids leakage between training and evaluation data. 

To distinguish interpolation within seen domains from transfer to unseen conditions, the evaluation protocol used in this paper is held-out volumes within seen datasets.
When cross-dataset transfer is reported, entire datasets were excluded from training and used only for testing.

\subsection{Model Architecture}
\label{sec:2.2sec}
SAMamba3D combines two encoders: a 3D-adapted SAM branch for boundary-sensitive representations and a hierarchical Mamba branch for volumetric context \cite{kirillov2023sam,gu2023mamba}. The two branches interact at selected depths, and a Feature-wise Linear Modulation (FiLM)-modulated multi-scale decoder produces the final \(K\)-class segmentation \cite{perez2018film}. The design is motivated by the fact that local boundary cues are necessary but insufficient in multiphase micro-CT, where ambiguous voxels often require broader geometric context for reliable classification.

\begin{figure}[h!]
  \centering
  \includegraphics[width=\linewidth, height=5cm]{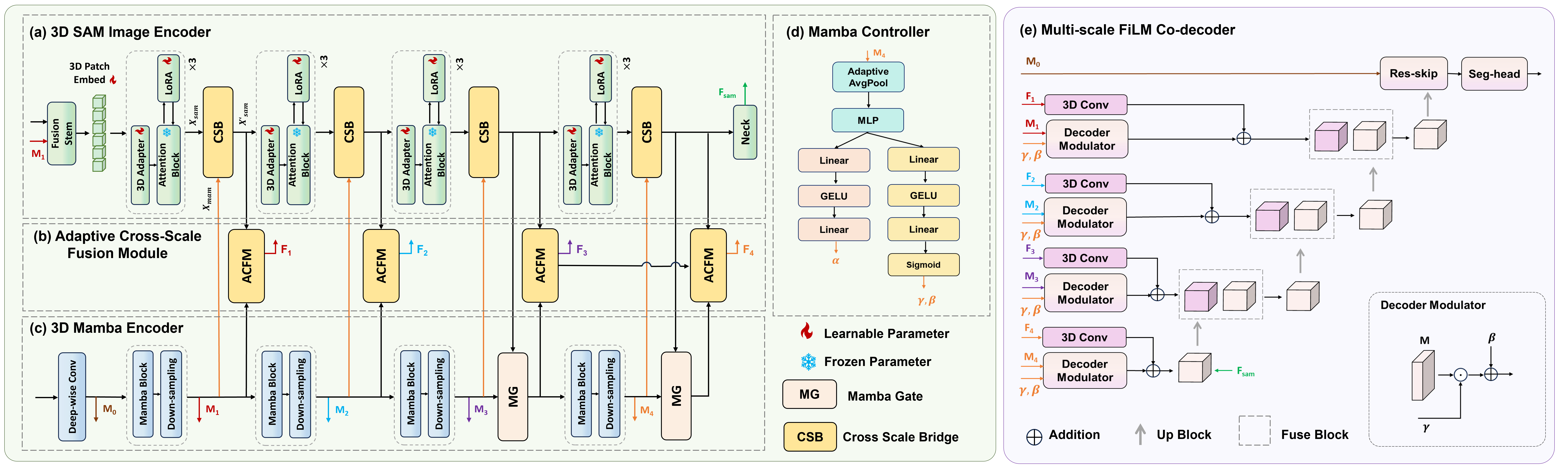}
  \caption{Architecture of SAMamba3D. (a) A 3D-adapted SAM image encoder extracts boundary-aware representations from volumetric inputs. (b) Adaptive cross-scale fusion modules, together with cross-scale bridges, enable bidirectional interaction between the SAM and Mamba branches across resolutions. (c) A hierarchical 3D Mamba encoder captures multi-scale volumetric context. (d) A Mamba controller generates the FiLM modulation parameters from Mamba features. (e) A multi-scale FiLM co-decoder progressively fuses and decodes the features to produce the final segmentation. }
  \label{fig:samamba3d_arch}
\end{figure}

As shown in Figure~\ref{fig:samamba3d_arch}, SAMamba3D comprises five tightly coupled modules: (a) a 3D-adapted SAM image encoder for boundary-aware representation learning; (b) a hierarchical 3D Mamba encoder for multi-scale volumetric context modeling; (c) adaptive cross-scale fusion modules and cross-scale bridges that enable bidirectional interaction between the SAM and Mamba branches; (d) a Mamba controller that transforms Mamba features into modulation parameters; and (e) a multi-scale FiLM co-decoder for progressive feature fusion and final segmentation.

\subsubsection{3D Adaptation of SAM}
Extending SAM from 2D natural images to volumetric pore-scale data requires changes in patch embedding, parameter adaptation, shallow feature injection, and positional encoding \cite{kirillov2023sam,ma2024medsam,wu2023medsamadapter,chen2024masam,yan2024aftersam}.

\paragraph{Fine-grained 3D patch embedding}
The original SAM encoder partitions a 2D image into \(16 \times 16\) patches. In 3D pore-scale imaging, coarse tokenization can suppress structures spanning only a few voxels, including narrow pore throats and thin wetting layers. We use a finer \(4 \times 4 \times 4\) 3D patch embedding:
\begin{equation}
T = \mathrm{Conv3D}(X;\; k=4,\, s=4)
\in
\mathbb{R}^{B \times \frac{D}{4} \times \frac{H}{4} \times \frac{W}{4} \times C},
\end{equation}
where \(X \in \mathbb{R}^{B \times 1 \times D \times H \times W}\) is the input volume, \(B\) is the batch size, \(D,H,W\) are spatial dimensions, and \(C=768\) is the feature dimension of the ViT-B backbone.

\paragraph{Parameter-efficient adaptation}
To reuse the pre-trained SAM representation without full retraining, we adopt low-rank adaptation (LoRA) \cite{hu2021lora}. For each adapted block, the residual update is
\begin{equation}
\Delta x = W_B \cdot \mathrm{Dropout}(W_A \cdot x) \cdot \frac{\alpha}{r},
\end{equation}
where \(x \in \mathbb{R}^{C}\), \(W_A \in \mathbb{R}^{C \times r}\), \(W_B \in \mathbb{R}^{r \times C}\), \(r=8\), and \(\alpha=16\). In the implementation used here, LoRA is inserted into attention projections.

\paragraph{Early volumetric fusion}
Before patch embedding, the shallowest Mamba feature map \(s_0\) is injected into the input pathway:
\begin{equation}
X' = X + |\lambda| \cdot \mathrm{Proj}(\mathrm{Upsample}(s_0)),
\end{equation}
where \(\lambda\) is a learnable scalar initialized to \(0.001\). This provides a high-resolution volumetric cue before tokenization and also preserves a direct pathway from early 3D features to the decoder.

\paragraph{Anisotropy-aware positional encoding}
Micro-CT data may be anisotropic, with different voxel spacing along axial and lateral directions. Instead of isotropically resampling the input volume, we encode anisotropy directly in the positional bias:
\begin{equation}
b(c, z) =
\sin\!\left(
\frac{2\pi(c+1)}{C}\cdot\frac{\rho z}{S}
\right),
\end{equation}
where \(c\) is the channel index, \(z\) is the depth coordinate, \(S\) is the base spatial scale, and \(\rho = \Delta z / \Delta x\) is the anisotropy ratio.

\subsubsection{Hierarchical Mamba Branch and Global Conditioning}
The Mamba branch performs hierarchical 3D feature extraction and provides both volumetric context and conditioning signals for feature interaction and decoding \cite{gu2023mamba}. It outputs multi-scale representations
\[
\{s_0, s_1, s_2, s_3\},
\]
with resolutions \(D/2\), \(D/4\), \(D/8\), and \(D/16\), and channel dimensions \([48,96,192,384]\), respectively.

\paragraph{Global descriptor}
The coarsest feature map \(s_3\) is converted into a compact global descriptor
\begin{equation}
G = \mathrm{MLP}(\mathrm{GAP}(s_3)) \in \mathbb{R}^{B \times d},
\end{equation}
where \(d=384\), \(\mathrm{GAP}(\cdot)\) denotes global average pooling, and \(\mathrm{MLP}(\cdot)\) is a two-layer perceptron with GeLU activation. The descriptor \(G\) is used to condition feature fusion and decoder modulation.

\paragraph{Spatial importance map}
To identify spatially informative regions, we derive an importance map from the intermediate feature map \(s_2\):
\begin{equation}
M = \sigma(\phi(s_2))
\in
[0,1]^{B \times 1 \times D' \times H' \times W'},
\end{equation}
where \(\phi(\cdot)\) is a lightweight prediction head and \(\sigma(\cdot)\) is the sigmoid function. In practice, this map assigns high scores to spatially informative regions, which are often concentrated near uncertain interfaces and boundary-rich areas.

\paragraph{Global conditioning signals}
From the global descriptor \(G\), we derive two sets of conditioning parameters. First, a vector of injection strengths \(\gamma \in (0,1)^K\) controls the contribution of global features at selected encoder depths. Second, FiLM parameters \((\gamma_j,\beta_j)\) are produced for each decoder stage \(j\):
\begin{equation}
\gamma = \sigma(\mathrm{MLP}_{\gamma}(G)),
\qquad
(\gamma_j,\beta_j) = \mathrm{MLP}_j(G).
\end{equation}
These parameters are volume-dependent and allow the model to modulate feature processing according to the global context of the current input.

\subsubsection{Layer-wise SAM--Mamba Interaction}
The two encoder branches interact at selected depths. Interaction is staged: early layers use one-way injection from the Mamba branch to the SAM branch, whereas deeper layers allow bidirectional exchange once the SAM features become semantically more stable.

\paragraph{Shallow one-way injection}
At early interaction depths, Mamba features are injected into the SAM branch through a dual-pathway adapter. The local pathway captures fine 3D structure using depthwise-separable 3D convolutions, and the global pathway aggregates coarse statistics through spatial pooling. The two pathways are combined using a learned gate:
\begin{equation}
F = w_l \cdot F_{\text{local}} + w_g \cdot F_{\text{global}},
\end{equation}
\begin{equation}
[w_l,\, w_g] =
\mathrm{Softmax}\!\Bigl(
W \cdot
[\mathrm{GAP}(F_{\text{local}}),\, F_{\text{global}}]
\Bigr),
\end{equation}
where \(F_{\text{local}} \in \mathbb{R}^{B \times C_m \times D_m \times H_m \times W_m}\), \(F_{\text{global}} \in \mathbb{R}^{B \times C_m}\), and \(W\) is a learned projection matrix. At these depths, reverse flow from the SAM branch to the Mamba branch is disabled.

\paragraph{Deep bidirectional interaction}
At deeper interaction depths, SAM features contain more stable boundary information and are projected back into the Mamba feature space. The reverse pathway is gated by a learnable scalar initialized to zero, so that feedback is introduced gradually during training rather than at initialization.

\paragraph{Selective computation}
The spatial importance map $M$ derived from the global controller enables differentiated processing within the SAM encoder. Regions with high importance scores, typically phase interfaces, are processed using full attention-based encoding, while low-information regions are handled through a lightweight pathway.
This routing strategy aligns computational effort with the spatial distribution of physically meaningful structures. Since informative regions occupy only a small fraction of the volume, this mechanism reduces redundant computation while preserving segmentation accuracy.

\paragraph{Cross-scale fusion}
At each interaction stage, SAM features, Mamba features, and cross-layer context are fused by
\begin{equation}
F_{\text{out}} =
\alpha \cdot \mathrm{Attn}(F_{\text{sam}})
+
\beta \cdot \mathrm{SE}(\mathrm{Proj}(F_{\text{mamba}}))
+
\gamma \cdot \mathrm{CA}(\mathrm{Proj}(F_{\text{ctx}})),
\end{equation}
where \(\alpha + \beta + \gamma = 1\) through softmax normalization. This prevents any single feature source from dominating the fused representation.

\subsubsection{FiLM-modulated Multi-scale Decoder}
The decoder reconstructs voxel-level predictions while combining global consistency with local boundary fidelity. Starting from the coarsest scale, it progressively upsamples and refines the feature representation using the Mamba skip features, the cross-encoder fusion features, the global conditioning signals, and the high-resolution stem feature.

\paragraph{FiLM modulation}
At decoder stage \(j\), the Mamba skip feature \(s_j\) is modulated by the global descriptor \cite{perez2018film}:
\begin{equation}
s'_j = \gamma_j(G) \odot s_j + \beta_j(G),
\end{equation}
where \(\gamma_j(G)\) and \(\beta_j(G)\) are channel-wise scaling and shifting terms predicted from \(G\). The parameters are initialized so that the modulation is initially close to the identity map.

\paragraph{High-resolution stem pathway}
To preserve thin interfaces and small structures, we reintroduce the high-resolution stem feature \(F_{\text{stem}}\) during reconstruction:
\begin{equation}
x' = x + \lambda \cdot \mathrm{Refine}([x,\,F_{\text{stem}}]),
\end{equation}
where \(\lambda\) is a learnable gate initialized to zero.

\paragraph{Output head}
The final prediction is produced by a \(3\times3\times3\) convolution followed by a \(1\times1\times1\) projection to the \(K\)-class output space.

\subsection{Boundary-aware Composite Loss}
\label{sec:2.3sec}
Multiphase micro-CT segmentation involves two coupled difficulties: strong class imbalance and intrinsic uncertainty near phase boundaries. In typical pore-scale volumes, the solid matrix occupies most voxels, whereas the most physically important structures, thin wetting layers, narrow throats, and interfacial regions, occupy only a small fraction of the volume. Manual annotations are also least reliable at these interfaces because gray-level transitions are gradual and different labels can be assigned by different methods.

To reduce the influence of uncertain boundary labels while preserving sensitivity to small structures, we use a boundary-aware composite loss with three components: weighted Dice loss, Tversky loss, and confidence-weighted focal loss.

\paragraph{Boundary confidence weighting}
Let \(y(v)\) denote the hard one-hot label at voxel \(v\), and let \(d(v)\) denote the Euclidean distance from \(v\) to the nearest class boundary in the manual annotation. We define a confidence weight
\begin{equation}
w(v) = \min\!\left(1,\frac{d(v)}{\delta}\right),
\end{equation}
where \(\delta = 0.3\). Voxels far from class boundaries receive weight \(1\), whereas weights decrease smoothly inside the boundary band.

\paragraph{Soft targets near boundaries}
Within the same boundary band, hard labels are replaced by softened targets \(\hat y(v)\) to reduce over-penalization of ambiguous interface voxels.
\begin{equation}
\hat y_c(v) =
\begin{cases}
y_c(v), & d(v)\ge \delta,\\[4pt]
(1-\eta(v))\,y_c(v) + \eta(v)\,\frac{1}{K}, & d(v)<\delta,
\end{cases}
\end{equation}
with
\begin{equation}
\eta(v) = 1-\frac{d(v)}{\delta}.
\end{equation}

\paragraph{Weighted Dice loss}
Class imbalance is handled using inverse-volume class weights
\begin{equation}
w_c = \frac{V_{\text{total}}}{K\cdot V_c},
\end{equation}
where
\begin{equation}
V_c = \sum_v \hat y_c(v),
\qquad
V_{\text{total}} = \sum_{c=1}^{K} V_c.
\end{equation}
The weighted Dice loss is
\begin{equation}
\mathcal{L}_{\text{Dice}} =
\frac{1}{K}
\sum_{c=1}^{K}
w_c \left(
1-
\frac{2\sum_v p_c(v)\hat y_c(v)+\varepsilon}
{\sum_v p_c(v)+\sum_v \hat y_c(v)+\varepsilon}
\right),
\end{equation}
where \(p_c(v)=\mathrm{softmax}(z(v))_c\) is the predicted probability for class \(c\).

\paragraph{Tversky loss}
To penalize missed small structures more strongly than spurious voxels, we use the tversky loss \cite{salehi2017tversky}
\begin{equation}
\mathcal{L}_{\text{Tversky}} =
\frac{1}{K}
\sum_{c=1}^{K}
\left(
1-
\frac{\mathrm{TP}_c+\varepsilon}
{\mathrm{TP}_c+\alpha \mathrm{FN}_c+\beta \mathrm{FP}_c+\varepsilon}
\right),
\end{equation}
with \(\alpha=0.7\) and \(\beta=0.3\). Here \(\mathrm{TP}_c\), \(\mathrm{FN}_c\), and \(\mathrm{FP}_c\) are computed using the softened targets \(\hat y_c(v)\).

\paragraph{Confidence-weighted focal loss}
To focus optimization on genuinely difficult voxels while reducing the contribution of uncertain boundary labels, we use \cite{lin2017focalloss}
\begin{equation}
\mathcal{L}_{\text{Focal}} =
\frac{\sum_v w(v)\,(1-p_t(v))^{\gamma_f}\,[-\log p_t(v)]}
{\sum_v w(v)},
\end{equation}
where \(p_t(v)\) is the predicted probability assigned to the target class at voxel \(v\), and \(\gamma_f=2\).

\paragraph{Total loss}
The total loss is a weighted sum:
\begin{equation}
\mathcal{L}
=
\lambda_1 \mathcal{L}_{\text{Dice}}
+
\lambda_2 \mathcal{L}_{\text{Tversky}}
+
\lambda_3 \mathcal{L}_{\text{Focal}},
\end{equation}
with \(\lambda_1=1.0\), \(\lambda_2=0.5\), and \(\lambda_3=0.5\). In this configuration, the Dice term provides the dominant overlap objective, while the tversky and focal terms act as auxiliary terms that improve sensitivity to small structures and reduce the influence of uncertain boundary labels.

\subsection{Implementation}
\label{sec:2.4sec}
Training a dual-encoder architecture containing both pre-trained and randomly initialized components is unstable if all parameters are optimized jointly from the start. We therefore used a two-stage progressive unfreezing strategy.

\paragraph{Stage A: 3D warm-up}
In the first stage, only the 3D-specific components were trainable: the Mamba branch, the early fusion stem, the 3D patch embedding, the forward adapters, the cross-encoder fusion modules, and the decoder. All SAM attention blocks, LoRA modules, and reverse feedback gates remained frozen. The reverse pathway from the SAM branch to the Mamba branch was disabled.

\paragraph{Stage B: SAM adaptation and reverse interaction}
Starting from the Stage A checkpoint, we unfroze the SAM layernorm layers, the 3D adapters, the LoRA modules, and the reverse feedback gates. This allows boundary-sensitive SAM features to adapt gradually to the volumetric domain while preserving the stability of the pretrained backbone.

The model was optimized using the AdamW optimizer with $\beta_1 = 0.9$, $\beta_2 = 0.999$, a weight decay of $10^{-5}$, and an initial learning rate of $10^{-4}$. Training was conducted for a maximum of 200 epochs, with early stopping applied based on the validation loss using a patience of 20 epochs.
The input patch size was set to $96 \times 96 \times 96$ with a batch size of 2. The training and testing was performed on a single NVIDIA A100 GPU with 80 GB memory.

We adopted a two-stage training schedule to stabilize optimization. In Stage A (warm-up phase), the model was trained for the first 30 epochs to gradually adapt the randomly initialized modules while preserving the integrity of pretrained representations. In Stage B, the full model was jointly optimized for the remaining epochs to achieve convergence and maximize segmentation performance.

\paragraph{Inference}
At test time, the full 3D volume was partitioned into overlapping patches. Predictions were combined using Gaussian-weighted stitching to suppress edge artifacts introduced by truncated receptive fields:
\begin{equation}
\hat{Y}(v) =
\frac{\sum_i G_i(v)\,P_i(v)}
{\sum_i G_i(v)},
\end{equation}
where \(P_i(v)\) is the softmax prediction of the \(i\)-th patch at voxel \(v\), and \(G_i(v)\) is the corresponding Gaussian weight. The overlap ratio was set to \(0.7\).

\subsection{Evaluation Metrics}
\label{sec:2.5sec}
We evaluated performance at two levels: (a) voxel-level segmentation overlap and (b) preservation of downstream physical descriptors computed from the segmented images.

\paragraph{Voxel-level overlap metrics}
All overlap metrics were computed per class and reported as macro-averaged values across classes.
The Dice similarity coefficient (DSC) is
\begin{equation}
\label{eq:dice}
\mathrm{DSC} = \frac{2|P \cap G|}{|P| + |G|},
\end{equation}
where \(P\) and \(G\) denote the predicted (the segmentation derived from our method) and reference (the segmentation for our test datasets as published in the literature) voxel sets for a given class.  

The intersection over union (IoU) is
\begin{equation}
\mathrm{IoU} = \frac{|P \cap G|}{|P \cup G|},
\end{equation}
where \(|P \cup G| = |P| + |G| - |P \cap G|\). DSC measures volumetric overlap, whereas IoU is stricter and more sensitive to both over-segmentation and under-segmentation.

\paragraph{Downstream physical properties}
Because the scientific relevance of segmentation depends on the preservation of pore-scale descriptors, we also evaluated the discrepancy between properties computed from predicted and reference segmentations \cite{scanziani2017contactangle,gao2024gradientseg,huang2021topology,ma2025superres}. The property set reported in this paper is porosity, saturation, interfacial area, Euler number, contact angle and curvature.
When contact angle and interfacial curvature are reported, they were computed using established pore-scale image-based measurement workflows rather than {\it ad hoc} geometric proxies \cite{andrew2014contact,alratrout2017automatic,lin2018curvature}.

This second level of evaluation is essential because small voxel-level errors can have disproportionately large effects on connectivity, interface geometry, and other quantities used in pore-scale interpretation.

\section{Results}
\label{sec:others}
\subsection{Comparison with Other Methods}
We benchmarked SAMamba3D against six representative 3D segmentation architectures that cover the dominant design paradigms in 3D image analysis: the canonical convolutional encoder–decoder (U-Net), the auto-configuring convolutional framework (nnU-Net \cite{isensee2021nnunet}), the pure-transformer volumetric network (UNETR), the residual-convolutional variant (SegResNet \cite{myronenko2018segresnet}), the shifted-window transformer (SwinUNETR \cite{hatamizadeh2022swinunetr}), and the hybrid state-space architecture (U-Mamba \cite{ma2024umamba}). All models were trained and evaluated under identical data splits, patch sizes, augmentation protocols, and optimization schedules, ensuring that observed differences reflect architectural inductive biases rather than training heuristics.

\begin{figure}[htbp]
  \centering
  \begin{minipage}{0.8\textwidth}
  \includegraphics[height=8.5cm]{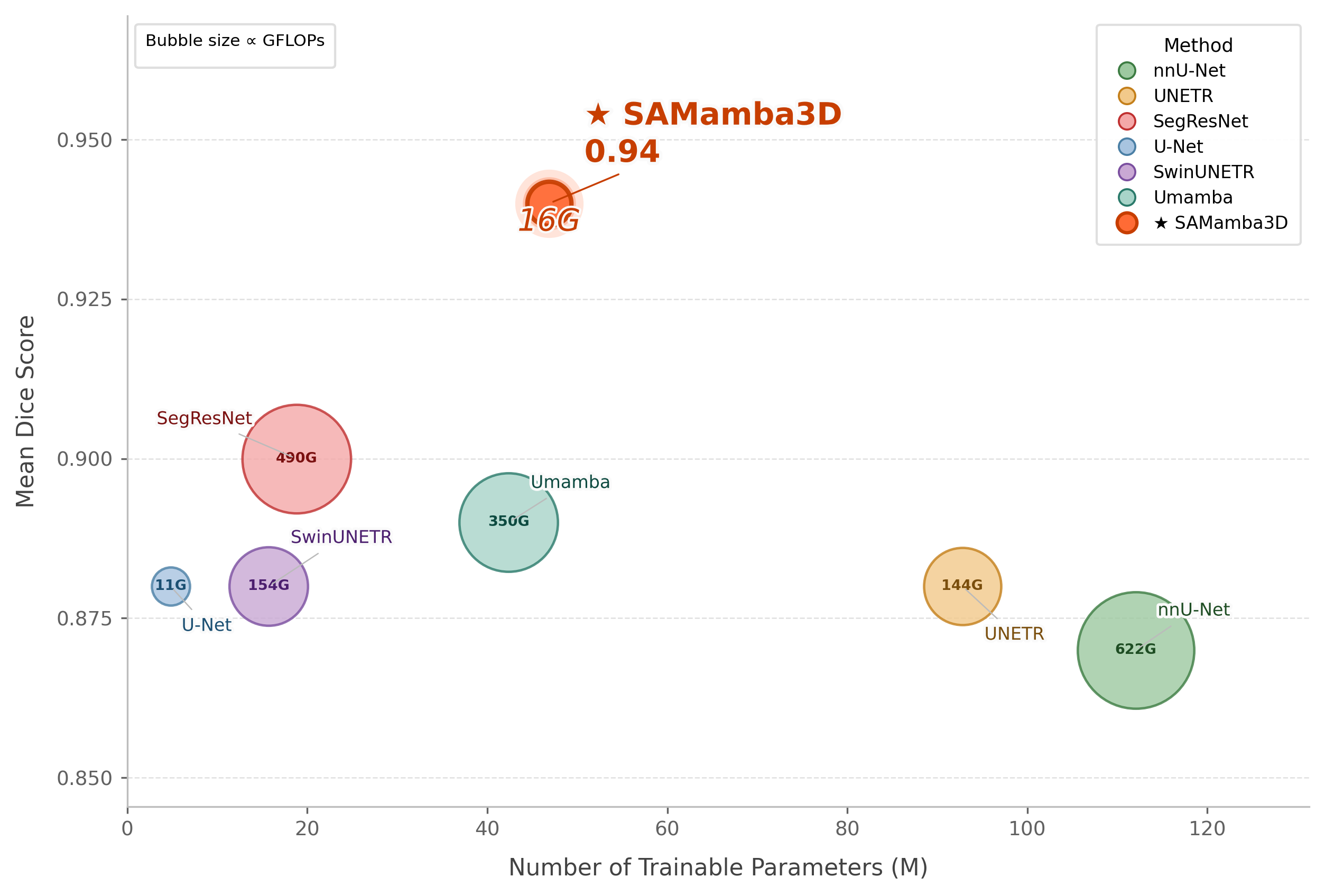}
  \caption{Accuracy–efficiency trade-off across the seven 3D segmentation models evaluated under identical training and validation protocols. The horizontal axis denotes the number of trainable parameters (in millions, M), the vertical axis denotes the mean Dice score computed on two datasets not used for training, and the marker size denotes the time needed to generate a new segmentation after training: it is the inference-time computational cost, measured in GFLOPs for a single forward pass with the same input volume. Each color corresponds to one method, as indicated in the legend. The orange star marks the proposed SAMamba3D in the upper-left region that is simultaneously more accurate and more lightweight.}
  \label{fig:fig2}
  \end{minipage}
\end{figure}

Figure~\ref{fig:fig2} compares segmentation accuracy, trainable parameter count, and computational cost. The segmentation accuracy is quantified by the Dice score defined in Eq.~(\ref{eq:dice}) computed as the average over two unseen test cases, Table~\ref{tab:table1}: Bentheimer sandstone under water-wet conditions~\cite{shojaei2023}, which is broadly consistent with the training domain; and Estaillades carbonate under oil-wet conditions~\cite{Hussain2025esta}, which represents a distinct rock type and wetting state. Both volumes were excluded from training and validation. The reference labels were obtained from the published segmentation. In Figure~\ref{fig:fig2}, the marker area represents the floating-point operations required to generate a new segementation on one of the test datasets. It represents the inference cost, measured in GFLOPs per forward pass for a \(96 \times 96 \times 96\) input patch. They reflect the relative deployment cost of different architectures rather than the total training cost. The horizontal axis gives the number of trainable parameters. 

SAMamba3D achieves the highest mean Dice score of 0.94, with 47.6 M trainable parameters and an inference cost of 16 GFLOPs per forward pass, demonstrating a favorable accuracy-efficiency trade-off.
Relative to nnU-Net, the long-standing reference for both medical and pore-scale segmentation \cite{isensee2021nnunet}, the proposed model reduces computational cost by nearly a factor of 40 (16 GFLOPS vs. 622 GFLOPs) while simultaneously improving the Dice score by 15\%. The comparison with U-Mamba is particularly informative: although both architectures incorporate state-space modeling, SAMamba3D achieves a 22-fold reduction in FLOPs with higher accuracy. This indicates that the performance gain does not arise from the use of Mamba alone, but from how global state-space context is integrated into a frozen, preTrained SAM backbone through the bidirectional bridge.

A notable aspect of this result is that the accuracy–efficiency frontier defined by the baseline models is not simply extended, but fundamentally shifted. SAMamba3D occupies a region of the trade-off space that is otherwise unpopulated. 

At the same time, transformer-based baselines (UNETR, SwinUNETR) and the largest convolutional model (nnU-Net) cluster within a narrow dice score range despite large differences in parameter count. This suggests that raw model capacity is not the limiting factor for micro-CT segmentation. Instead, performance is governed by the ability to reconcile local voxel-level evidence with global pore-network topology—a requirement that is directly addressed by coupling a vision foundation model with a linear-complexity sequence model.

\subsection{Image Analysis}
\paragraph{Water-wet Bentheimer sandstone}
We further examined out-of-distribution performance on an unseen water-wet Bentheimer sandstone dataset. This dataset was not used during training or validation and was selected as a near-domain benchmark because it represents a classical oil--brine segmentation problem in sandstone. The reference labels were obtained from the published benchmark segmentation provided with the dataset~\cite{shojaei2023}.
\begin{figure}[h]
  \centering
  \includegraphics[width=\textwidth]{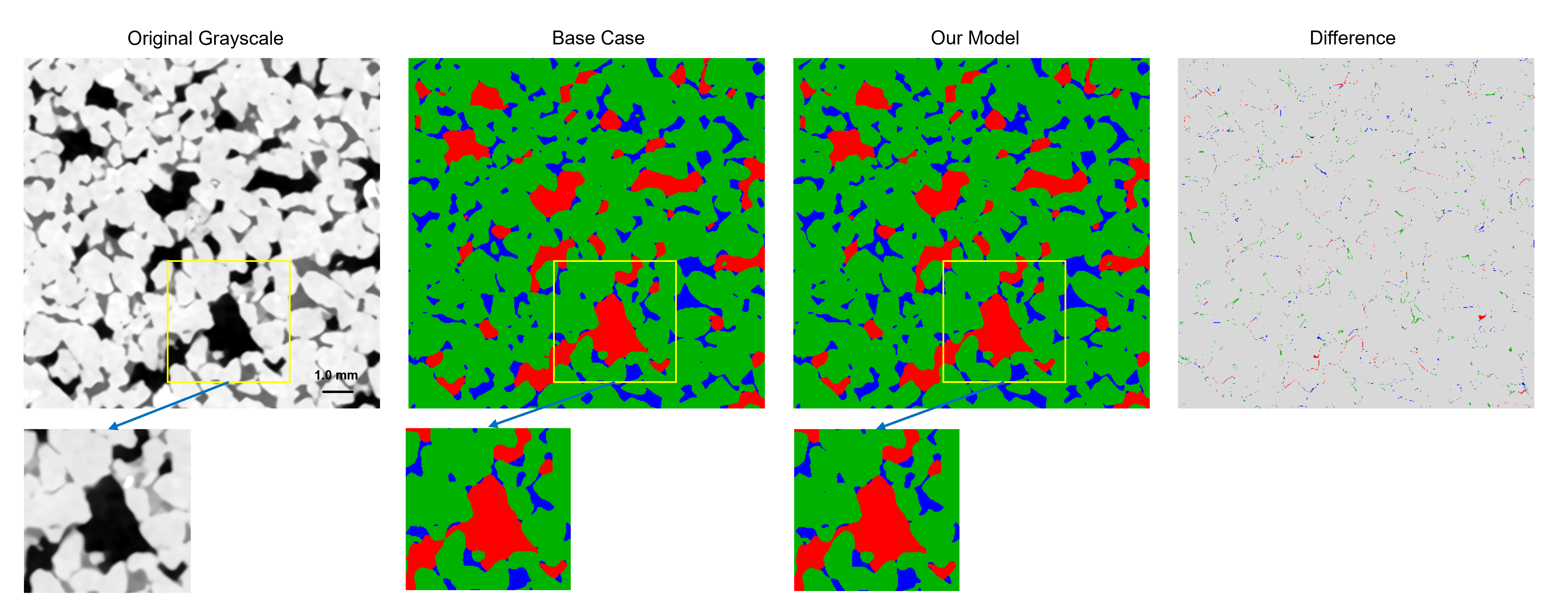}
  \caption{Model performance on test set (Bentheimer, water-wet, see Table~\ref{tab:table1}). A representative 2D cross-section of a 3D image is shown, illustrating the original grayscale image (left), the base case segmentation and the results of our model SAMamba3D (middle), and the difference between them (right). In the segmented images, green is rock, brine is blue and oil is red. The yellow box highlights differences in the segmentation: our method captures better connected wetting layers of brine.}
  \label{fig:waterwet-ben}
\end{figure}

\begin{table}[htbp]
    \centering
    \begin{threeparttable}
    \caption{Comparison of baseline segmentation and SAMamba3D results based on the physical properties of water-wet Bentheimer sandstone with $512 \times 512 \times 512$ voxels. }
    \label{tab:properties}
    \begin{tabular}{lcc}
        \toprule
        \textbf{Property} & \textbf{Base case} & \textbf{Our model} \\
        \midrule
        Porosity (\%)                                                  & 20.30   & 21.02  \\
        Saturation of brine, $S_w$ (\%)                                & 47.10   & 49.64  \\
        Saturation of oil, $S_o$ (\%)                                  & 52.90   & 50.36  \\
        Interfacial area oil--brine ($\times 10^{6}\ \mu\mathrm{m}^2$) & 2.96    & 2.80   \\
        Interfacial area brine--grains ($\times 10^{6}\ \mu\mathrm{m}^2$) & 20.97 & 23.08 \\
        Interfacial area oil--grains ($\times 10^{6}\ \mu\mathrm{m}^2$)   & 7.82  & 7.91  \\
        Surface area of grains ($\times 10^{6}\ \mu\mathrm{m}^2$)      & 31.91   & 34.08  \\
        Euler number of pore space                        & -2796 & -3815 \\
       Euler number of brine                                & 1931 & -1384 \\
        Euler number of oil                                & 342  & 290 \\
        \bottomrule
    \end{tabular}
  \end{threeparttable}
\end{table}

As shown in Figure~\ref{fig:waterwet-ben}, SAMamba3D and the base-case segmentation agreed across the bulk of the pore space, but diverged systematically along narrow brine layers, sub-resolution throats, and phase contacts within large pore bodies. The zoomed region (yellow box) highlights a cluster in which the base-case segmentation assigns a large connected region to oil, whereas our model recovers thin wetting layers of brine that drape along grain surfaces and partition the oil ganglion into two physically distinct bodies. This is expected strongly water-wet behavior: in capillary equilibrium, the wetting phase remains topologically connected through corner filaments and wetting layers in the angular pore space \cite{gao2020watermixedwet,alhosani2021wettability}. The difference map shows that discrepancies concentrate on phase boundaries and isolated sub-voxel features rather than on bulk regions, ruling out a systematic bias in the prediction.
Quantitative physical properties computed on a volume of $512 \times 512 \times 512$ (see in Table ~\ref{tab:properties}) make this interpretation precise.

The Euler characteristic of each phase provides the most diagnostic comparison between the two segmentations. A positive Euler number indicates a phase that is fragmented into many disconnected components, whereas a negative value is consistent with a well-connected, multiply-handled network, which is the topology typically expected for the wetting phase in a water-wet sandstone at intermediate saturation \cite{huang2021topology,gao2020wwmw,lin2019minimal}. Our prediction yields an Euler number for brine of –1384, compared with 1931 for the base case, Table~\ref{tab:properties}. The Euler number of the pore space also decreases from –2796 to –3815, which is consistent with the model capturing additional small-scale features that add loops to the network rather than introducing isolated noise voxels, as the latter would raise the Euler number rather than lower it.

These observations together indicate that the improvement over the base case is not limited to the saturation: the predicted brine phase exhibits a connectivity that is more consistent with the wetting state of the sample, whereas the base-case segmentation yields a brine topology that is more difficult to reconcile with a continuously percolating wetting phase. Generating the segmented image on a $512^{3}$ volume took approximately 8 minutes on a single A100 GPU, which provides a reliable initial segmentation and reduces the need for the hours of manual correction that such volumes typically require.

\paragraph{Mixed-wet Bentheimer sandstone}
A more demanding test is provided by the mixed-wet Bentheimer sandstone of Lin et al. \cite{lin2019bentheimer}, in which wettability heterogeneity gives rise to complex fluid configurations that cannot be predicted by gray-level thresholding alone. In such systems the fluid morphology encodes local wettability, so the segmentation must faithfully reconstruct the geometry of the interface, not just the phase labels, if contact-angle and curvature measurements are to remain physically meaningful.
\begin{figure}[h]
  \centering
  \includegraphics[width=\textwidth]{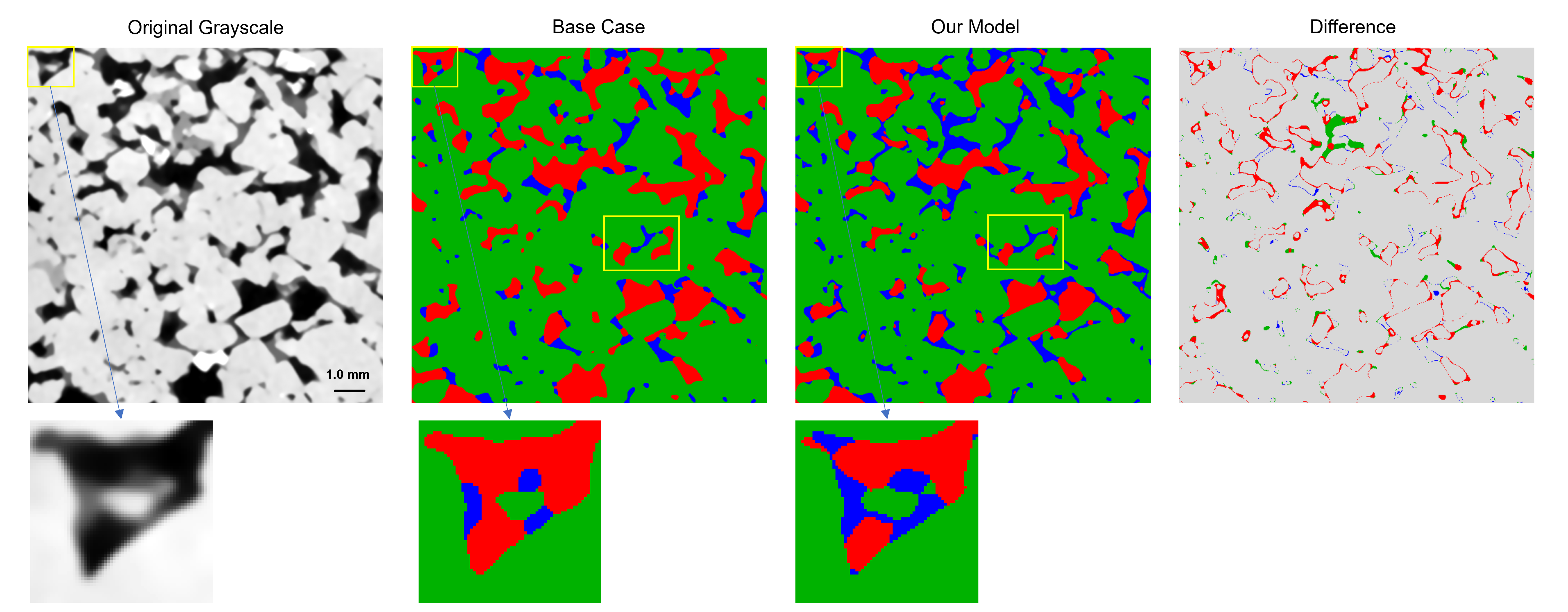}
  \caption{Model performance on a test dataset (Bentheimer, mixed-wet, see Table~\ref{tab:table1}). A representative 2D cross-section of a 3D image is shown, illustrating the original grayscale image (left), the base case segmentation and the results of our model SAMamba3D (middle), and the difference between them (right). In the segmented images, green is rock, brine is blue and oil is red. The highlighted regions (yellow box) illustrate differences in the segmentation.}
  \label{fig:mixedwet-ben}
\end{figure}
Figure ~\ref{fig:mixedwet-ben} compares the grayscale input, the base-case segmentation, the SAMamba3D prediction, and their difference. The enlarged region (yellow box) is particularly informative. In the base case, the pore space shown is filled by a connected region of oil interrupted by a small brine cluster. By contrast, SAMamba3D resolves a more structured arrangement in which brine occupies opposing pore corners while oil remains in the center of the pore space. This geometry is consistent with mixed-wet pore occupancy, in which local wetting preferences vary along the grain surface and the resulting interface morphology is correspondingly heterogeneous. Importantly, the difference map shows that most discrepancies are concentrated along phase boundaries, with comparatively few discrepancies in the center of oil-filled or brine-filled regions. The improvement is therefore primarily an interfacial refinement, rather than a wholesale reassignment of phase volumes. 

\paragraph{Oil-wet Estaillades carbonate}
The oil-wet Estaillades carbonate \cite{Hussain2025esta} provides a distinct cross-domain test, as both its pore architecture and wetting condition differ markedly from the sandstone cases. Carbonates typically contain irregular pore shapes, rough grain surfaces, micro-porous regions, and locally variable gray-level contrast, all of which increase the ambiguity of phase assignment. Under oil-wet conditions, the oil phase is expected to be preferentially associated with mineral surfaces, pore corners, and narrow crevices.

\begin{figure}[h]
  \centering
  \includegraphics[width=\textwidth]{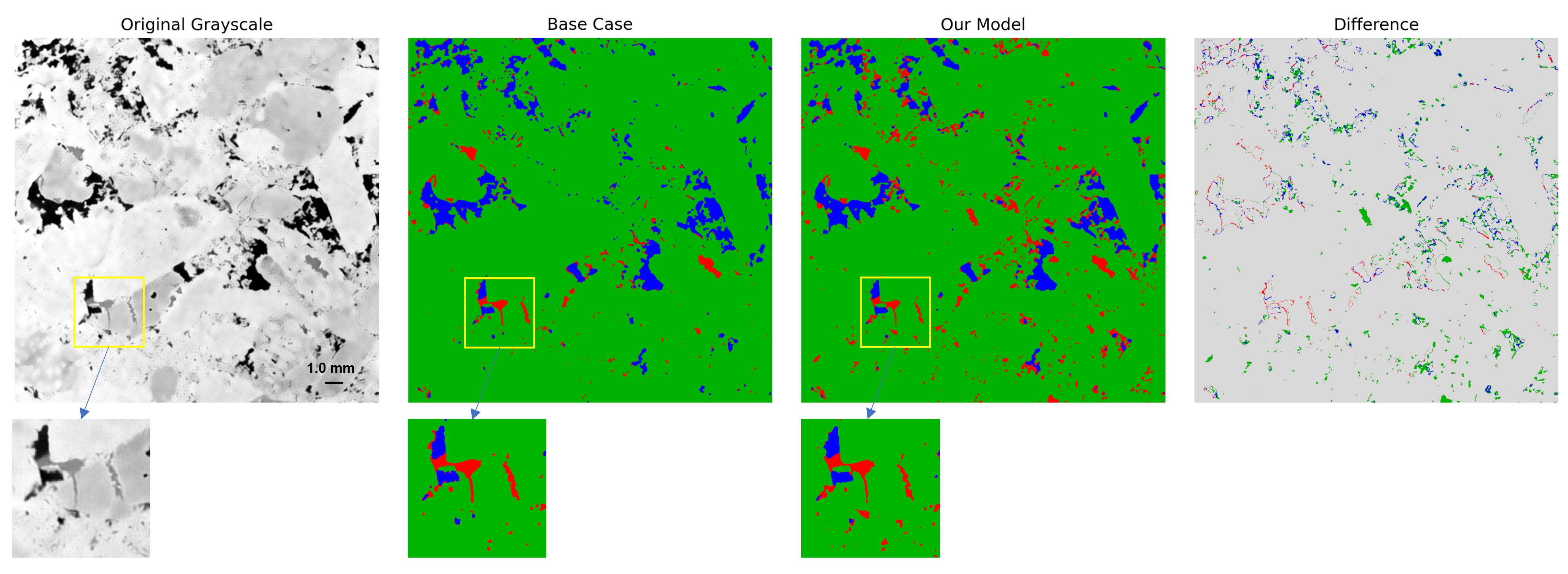}
  \caption{Model performance on a test dataset (Estaillades carbonate, oil-wet, see Table~\ref{tab:table1}). A representative 2D cross-section of a 3D image is shown, illustrating the original grayscale image (left), the base case segmentation and the results of our model SAMamba3D (middle), and the difference between them (right). In the segmented images, green is rock, brine is blue and oil is red. The highlighted regions (yellow box) show local differences in fluid morphology and interfacial reconstruction.}
  \label{fig:oilwet-esta}
\end{figure}

Figure~\ref{fig:oilwet-esta} suggests that SAMamba3D produces a fluid configuration that is qualitatively consistent with the reported oil-wet condition. The highlighted region (yellow box) provides a representative example. In the base case, several narrow interfacial features near the rock boundary are represented less distinctly. SAMamba3D predicts a more structured morphology, in which additional oil-bearing voxels are assigned to wall-adjacent and corner regions, while brine occupies the neighboring pore-body space. This arrangement is compatible with oil-wet pore occupancy, where the oil phase may remain in contact with the solid surface and angular pore regions. This observation should be interpreted as morphological consistency rather than independent proof of improved wetting-phase topology.

The difference map shows that most discrepancies are localized at rock--fluid and fluid--fluid interfaces, as well as in small pore features where the gray-level contrast is weak. By contrast, the bulk rock phase remains largely unchanged. This pattern indicates that the changes introduced by SAMamba3D are primarily interfacial refinements rather than a wholesale reassignment of phase volumes. Such refinement is particularly important for oil-wet carbonates, where contact-angle measurements, interfacial curvature, and estimates of wetting-phase contact with the solid are highly sensitive to errors at the solid boundary. These results suggest that SAMamba3D combines local gray-level evidence with three-dimensional structural context to refine ambiguous interfacial voxels in a manner compatible with the reported oil-wet pore-scale morphology.

\paragraph{Curvature and Contact Angle}
The geometry of the interface between the two fluid phases can be quantified by the distributions of the principal curvatures \(\kappa_1\) and \(\kappa_2\) on the fluid--fluid interface and the distribution of \textit{in situ} contact angles measured along the three-phase contact loops. These descriptors probe different but related aspects of the segmentation: contact-angle statistics reflect local wetting heterogeneity, whereas curvature statistics reflect the geometric state of the interface and can be used to quantify capillary pressure \cite{armstrong2012linking}.

\begin{figure}[h!]
    \centering
    \begin{subfigure}[b]{0.48\textwidth}
        \centering
        \includegraphics[width=\linewidth]{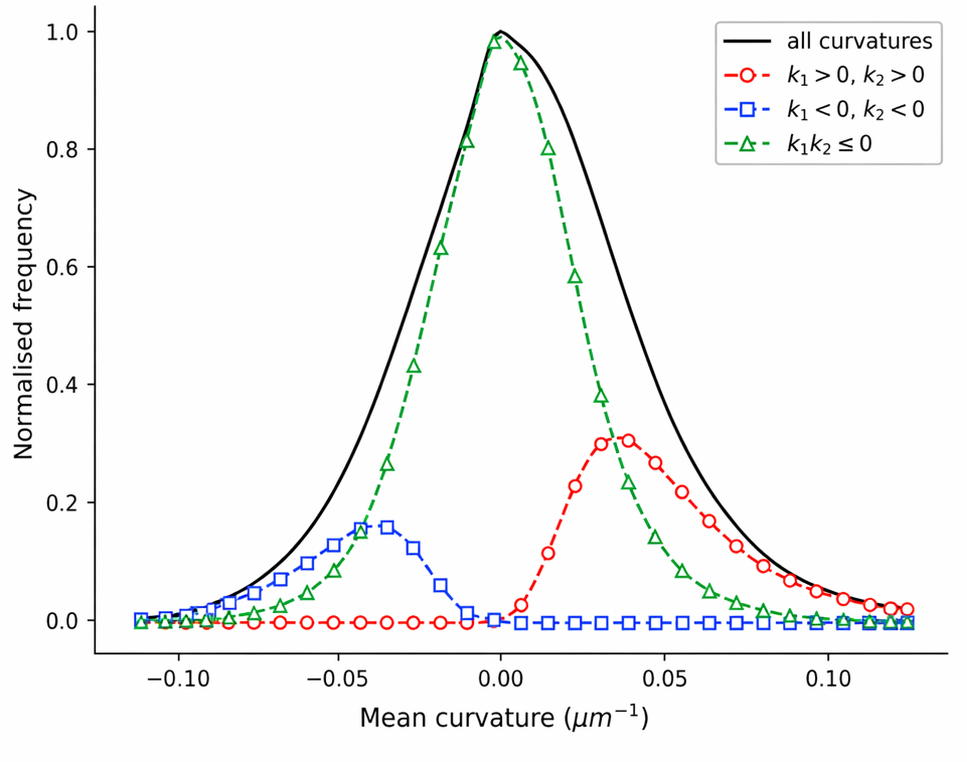}
        \caption{Our model (mixed-wet)}
        \label{fig:sub_a}
    \end{subfigure}
    \hfill
    \begin{subfigure}[b]{0.48\textwidth}
        \centering
        \includegraphics[width=\linewidth]{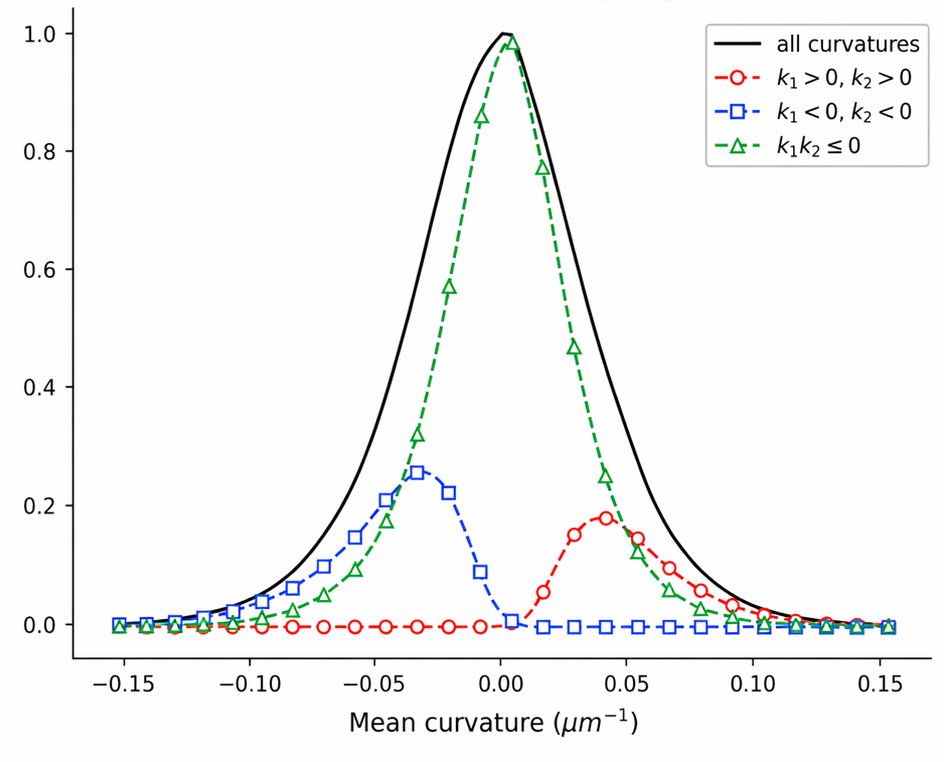}
        \caption{Base case (mixed-wet)}
        \label{fig:sub_b}
    \end{subfigure}
    \caption{Comparison of the distribution of measured interfacial curvature for the oil--brine interface reconstructed by SAMamba3D and the base-case segmentation. (a) SAMamba3D. (b) Base case. }
    \label{fig:curvature}
\end{figure}

\begin{figure}[h!]
    \centering
    \begin{subfigure}[b]{0.48\textwidth}
        \centering
        \includegraphics[width=\linewidth]{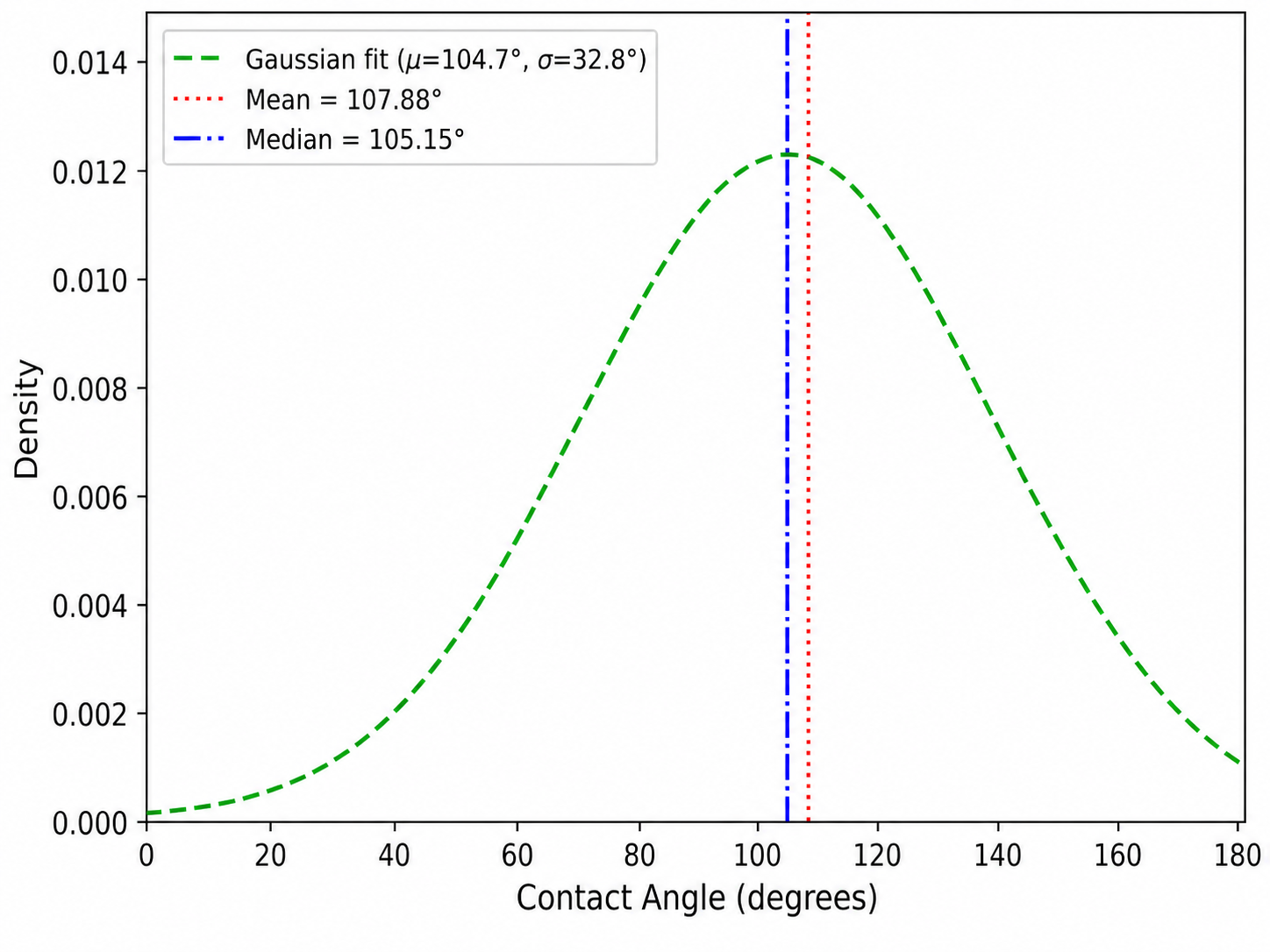}
        \caption{Our model (mixed-wet)}
        \label{fig:sub_a}
    \end{subfigure}
    \hfill
    \begin{subfigure}[b]{0.48\textwidth}
        \centering
        \includegraphics[width=\linewidth]{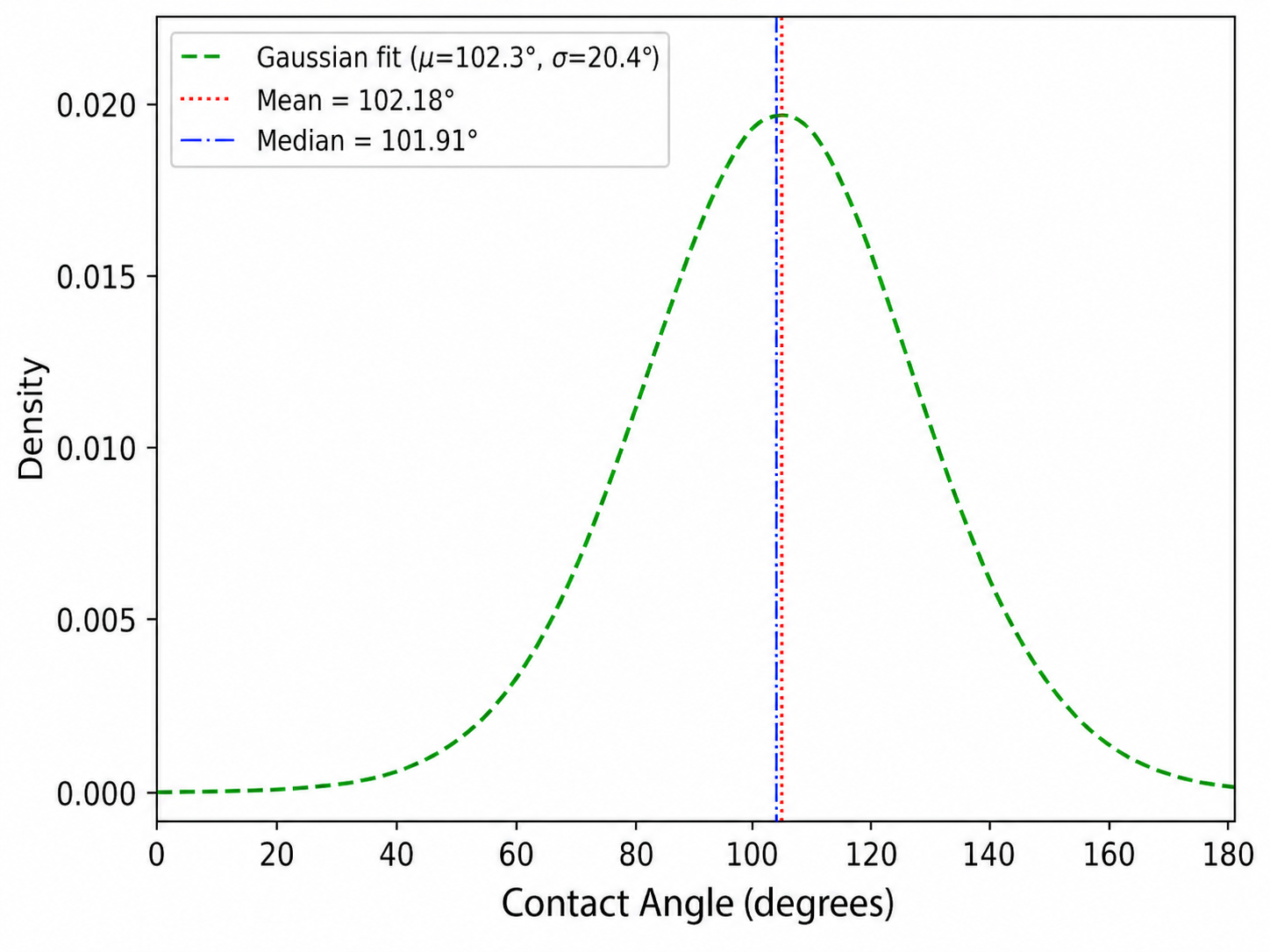}
        \caption{Base case (mixed-wet)}
        \label{fig:sub_b}
    \end{subfigure}
    \caption{Comparison of \textit{in situ} contact angle distributions for the mixed-wet sample reconstructed by SAMamba3D and the base-case segmentation. 
    (a) SAMamba3D. 
    (b) Base case.}
    \label{fig:contactangle}
\end{figure}

Figure~\ref{fig:curvature} decomposes the distribution of curvature into three: convex (\(\kappa_1>0\) and \(\kappa_2>0\)), concave (\(\kappa_1<0\) and \(\kappa_2<0\)), and saddle-like (\(\kappa_1\kappa_2<0\)). Relative to the base case, SAMamba3D produces a larger saddle-like population. More importantly, the corresponding mean curvature \(H=(\kappa_1+\kappa_2)/ 2\) is more tightly concentrated around zero in these regions. These two observations are significant when taken together: opposite-signed principal curvatures indicate locally saddle-shaped interface elements, while a mean curvature close to zero indicates that these elements are closer to locally minimal-surface-like configurations. Such geometry is expected for interfaces approaching capillary equilibrium in a mixed-wet porous medium \cite{lin2019minimal}.

The contact-angle distributions, Figure ~\ref{fig:contactangle}, are similar for the two segmentations: the mean contact angle is \(102.3^\circ\) for the base case and \(104.7^\circ\) for SAMamba3D. This indicates that both segmentations recover the overall mixed-wet character of the sample. The more informative difference lies in the spread of the distributions. SAMamba3D yields a substantially broader distribution, with a standard deviation of \(32.8^\circ\) compared with \(20.4^\circ\) for the base case. This broader spread suggests that SAMamba3D preserves more of the local angular variability associated with mixed-wet pore surfaces, whereas the base case appears to smooth away part of that variability.

The visual, statistical, and geometric evidence support a consistent interpretation. SAMamba3D does not merely reproduce the overall phase proportions of the base case. Rather, it recovers an interface geometry that is more compatible with the mixed-wet nature of the sample. This is the relevant criterion if the segmentation is to be used as a trustworthy input for contact-angle measurement, capillary-pressure interpretation, pore-network extraction, and direct numerical simulation of multiphase flow.

\section{Generalization}
We further evaluated SAMamba3D on a diverse collection of datasets that were not seen during training, with no retraining, no fine-tuning, and no per-case hyperparameter adjustment. Table~\ref{tab:table1} lists the full benchmark: six sandstones (Bentheimer with three independent acquisitions, Mt. Simon sandstone, borosilicate glass bead pack, and a layered Clashach sandstone) and seven carbonates (Middle East, Ketton, Estaillades, Indiana, and additional Middle Eastern samples), spanning water-wet, mixed-wet, and oil-wet conditions \cite{tawfik2022denoising,dalton2018methods,jangda2024subsurface,lin2018curvature,lin2019bentheimer,shojaei2023,Hussain2025esta,scanziani2018threephase,alhammadi2017mixedwet}. 
The associated fluid systems include oil–brine, sc\ce{CO2}–brine, and \ce{H2}–brine, which are the three fluid pairs of greatest relevance for subsurface energy applications: hydrocarbon recovery, geologic \ce{CO2} storage, and subsurface hydrogen storage.
\subsection{Different Rock Types and Wettability}
\begin{figure}[h!]
  \centering
  \includegraphics[width=\textwidth]{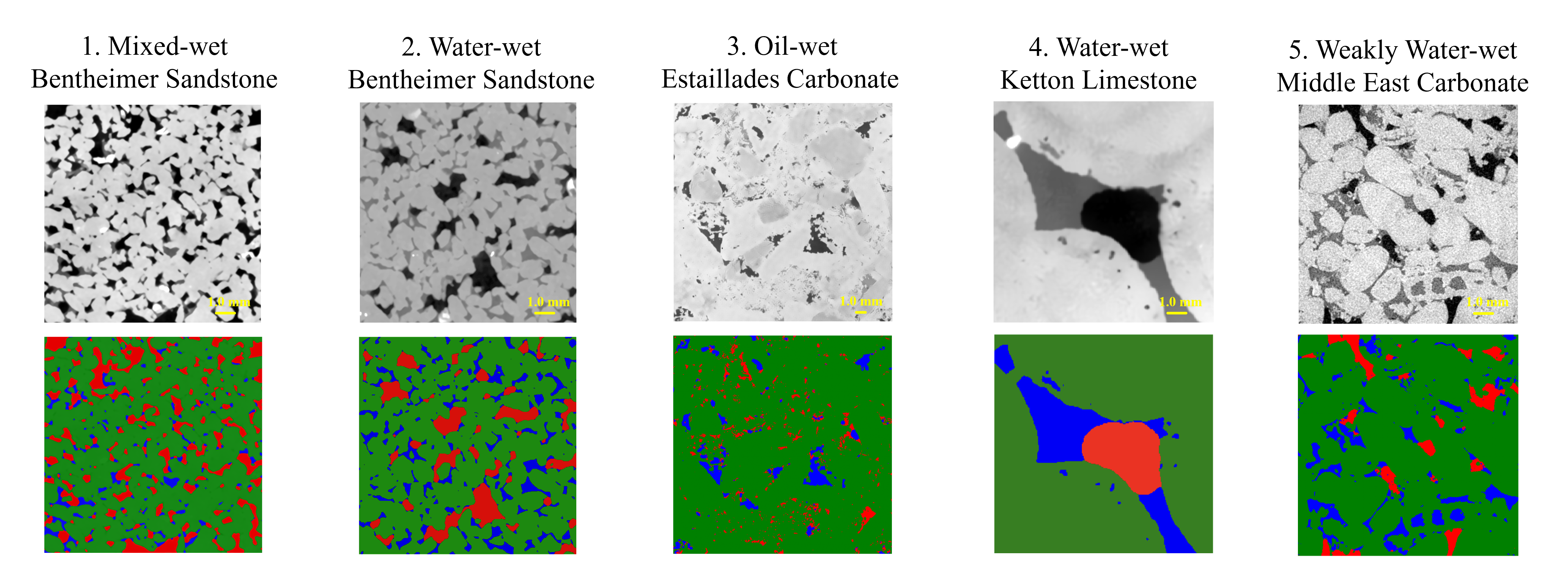}
  \caption{Generalization of SAMamba3D across diverse rock types and wetting conditions. Segmentation results are shown on five unseen datasets processed with a single set of trained weights and no fine-tuning. Top row: raw grayscale micro-CT cross-sections of 3D images. Bottom row: corresponding segmentation results from SAMamba3D, with green denoting the rock matrix, blue the denser fluid phase (brine), and red the less dense phase (oil).}
  \label{fig:rocktypes}
\end{figure}
Figure~\ref{fig:rocktypes} presents segmentations for five representative cases that together span the rock–wettability plane: (1) mixed-wet Bentheimer sandstone \cite{lin2019bentheimer}, (2) water-wet Bentheimer sandstone \cite{shojaei2023}, (3) oil-wet Estaillades carbonate \cite{Hussain2025esta}, (4) water-wet Ketton limestone \cite{scanziani2018threephase}, and (5) a weakly water-wet Middle Eastern carbonate \cite{alhammadi2017mixedwet}. These samples differ substantially in pore geometry, ranging from the well-sorted, intergranular porosity of Bentheimer, through the vulgar macro-porosity of Ketton, to the heterogeneous vug–matrix structure of the Middle Eastern carbonate. They also differ in grayscale contrast from one acquisition to the next. Despite this variability, SAMamba3D consistently delivered segmentations in which the three phases were spatially coherent, grain boundaries were sharp, and small pores were preserved.

Several morphologies compatible with the reported wetting states were observed without case-specific fine-tuning. In water-wet cases, the brine phase was preferentially assigned to grain-adjacent regions and concave pore corners, while oil was mainly located in pore-body regions. In the oil-wet Estaillades case, the predicted morphology showed the opposite tendency, with oil preferentially assigned to wall-adjacent and corner regions, while brine was more frequently located in neighboring pore-body regions. In the mixed-wet Bentheimer and weakly water-wet Middle Eastern carbonate, the two configurations coexisted within the same volume, and the model transitioned smoothly between them without discontinuities at domain boundaries. The emergence of these wettability-compatible morphologies from a single set of weights suggests that the model has learned transferable representations of interface geometry rather than relying only on texture statistics specific to any one sample.

Carbonate cases are particularly challenging because of their low phase contrast and the micro-porosity typical of such rocks. The Ketton limestone example, which contains large vulgar pores bounded by a micro-porous matrix, was segmented without the characteristic failure modes of threshold-based and 2D CNN methods, namely bleeding of the matrix phase into pore centers, or fragmentation of vug interiors \cite{gao2024gradientseg}. The weakly water-wet Middle East carbonate likewise retained a continuous brine layer that respected the topology expected from its wetting state, even though the grayscale contrast between micro-porous matrix and pore space is small. These are the cases in which conventional pipelines most often require extensive manual correction, and their successful handling by a single pretrained model is one of the strongest pieces of evidence for the generalization capacity of the proposed architecture. 

\subsection{Different Fluids}
\begin{figure}[h!]
  \centering
  \includegraphics[width=\textwidth]{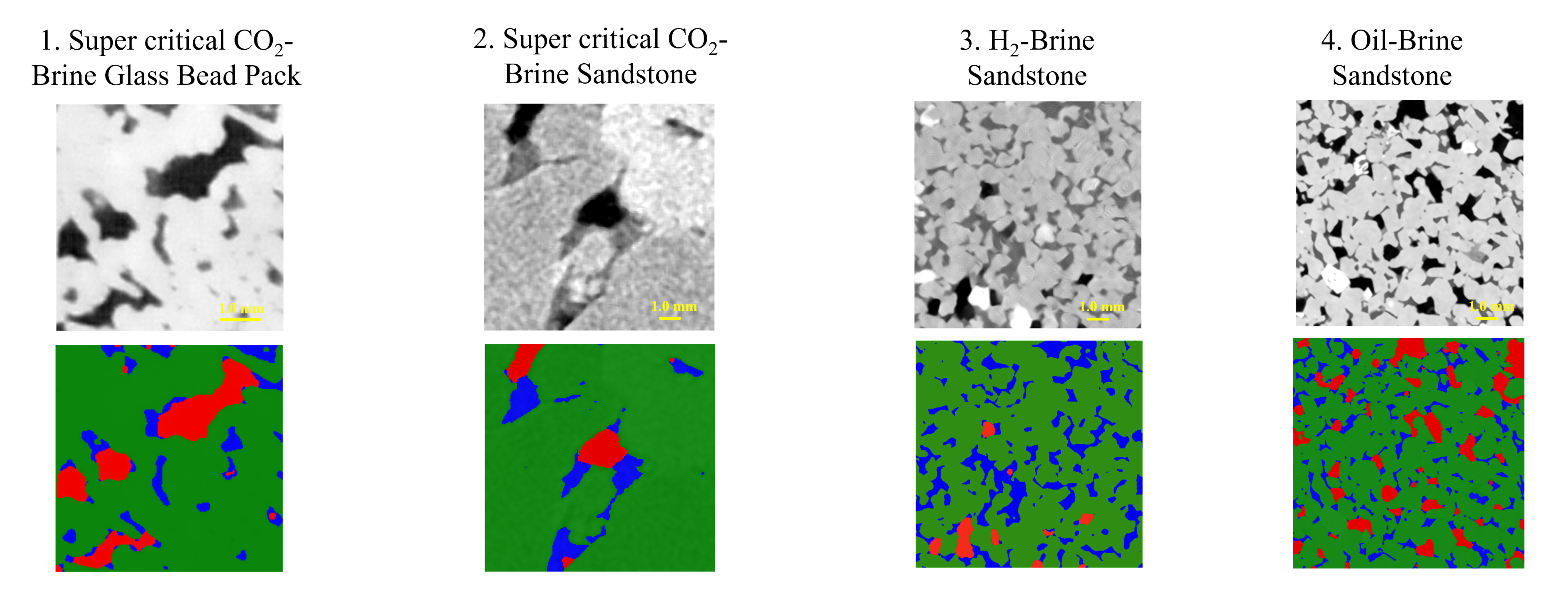}
  \caption{Generalization of SAMamba3D across different fluid systems. 
  Segmentation results on four unseen datasets that cover fluid pairs progressively distant from the oil–brine training distribution. Top row: raw grayscale micro-CT cross-sections. Bottom row: corresponding segmentation results from SAMamba3D, with green denoting the rock or bead matrix, blue the denser wetting phase (brine), and red the less dense non-wetting phase (sc\ce{CO2}, \ce{H2}, or oil).}
  \label{fig:fluid systems}
\end{figure}

Figure~\ref{fig:fluid systems} extends the analysis to fluid patterns that are increasingly distant from the oil–brine training distribution. Four representative cases are shown: (1) sc\ce{CO2}–brine in a glass bead pack \cite{tawfik2022denoising}, (2) sc\ce{CO2}–brine in a sandstone \cite{dalton2018methods}, (3) \ce{H2}–brine in a sandstone \cite{jangda2024subsurface}, and (4) oil–brine in a sandstone \cite{lin2018curvature}. The grayscale statistics of these systems differ substantially from the training data—sc\ce{CO2} and \ce{H2} exhibit grayscale intensities that overlap heavily with both brine and grain phases because of their low X-ray attenuation and the limited contrast afforded by typical laboratory micro-CT sources, yet SAMamba3D recovered the correct configurations throughout.

In the glass bead pack, the spherical grain geometry and idealized pore shapes produced a segmentation in which the non-wetting \ce{CO2} clusters formed well-defined ganglia with smooth, convex interfaces consistent with capillary-controlled drainage. In the sc\ce{CO2}–brine sandstone case, the more complex pore geometry was segmented with connected brine layers draping intragranular surfaces. In the \ce{H2}–brine sandstone case, where \ce{H2} images were not considered in the training data, the results are still physically consistent: \ce{H2} ganglia occupied pore centers, brine wrapped grains, and the grain phase was delineated without spurious speckles.

\section{Discussion}
The central result of this study is that multiphase pore-scale micro-CT segmentation can be treated as a reusable foundational model adaptation problem rather than applying dataset-specific training. For the cases considered here, a single SAMamba3D model transferred to new datasets without case-specific retraining while maintaining strong agreement not only in voxel-level overlap metrics, but also in the derived pore-scale descriptors evaluated in this study. This point is more important than a small numerical gain on a single benchmark, because in practical pore-scale workflows the main bottleneck is rarely whether a model can fit one curated dataset, but whether it remains reliable when rock type, wettability state, fluid system, or scanner characteristics change. In this respect, the present results suggest that reusable segmentation is achievable within a common multiphase micro-CT setting.

The observed transfer behavior is consistent with the design of the model. The adapted SAM branch contributes boundary-sensitive priors that would be difficult to learn robustly from limited pore-scale annotations alone, while the Mamba branch provides the 3D context needed to resolve ambiguous voxels and preserve 3D continuity. The multi-source training corpus and boundary-aware loss further reduce sensitivity to scanner-dependent gray-level statistics and uncertain labels near interfaces. Taken together, these components help explain why the model remains effective across the conditions studied, rather than collapsing to a narrowly sample-specific solution. Just as importantly, the computational cost remains substantially lower than that of fully retrained 3D baselines, which is relevant for full-volume deployment rather than patch-level benchmarking alone.

In the datasets examined here, the predicted segmentations preserved the pore-scale descriptors assessed in this study closely enough to serve as inputs for subsequent physical analysis. This is particularly relevant for multiphase imaging workflows in which manual correction is expensive and repeated retraining for each new experiment is impractical. More broadly, the results indicate that foundation-model adaptation can be a practical route for 3D images when local boundary cues must be combined with broader structural context \cite{zhu2024ipwgan,zhu2025diffusion}.

The present study also has important limitations. First, the training dataset remains dominated by sandstones and carbonates, and performance may therefore degrade in underrepresented lithologies such as organic-rich shales, highly fractured rocks, or unconsolidated granular media. Second, the model does not encode explicit physical constraints such as Young--Laplace consistency, minimal-surface regularization, or temporal continuity. Third, the current framework is developed for a single static image segmented into solid and two fluid phases; extending it to cases with more phases and time-resolved 4D imaging will require further work on data curation, supervision, and evaluation. These limitations point to clear directions for future research: broader and more balanced training data, physics-informed regularization, uncertainty estimation, and extension to temporally resolved pore-scale segmentation.

\section{Conclusions}
We presented SAMamba3D, a 3D segmentation framework that recasts multiphase pore-scale micro-CT segmentation as a foundation-model adaptation problem. By applying a largely frozen SAM encoder to volumetric data and coupling it with Mamba-based global context modeling, the method combines boundary-sensitive representations with 3D structural reasoning in a single segmentation pipeline.

Across the datasets and evaluation settings studied here, SAMamba3D outperformed the baselines considered while requiring substantially less computation than fully retrained 3D alternatives. More importantly, the model transferred across unseen data without case-specific retraining and preserved the downstream pore-scale descriptors, such as contact angle and curvature, evaluated in this study closely enough to support subsequent physical analysis. These results suggest that reusable foundation-model adaptation is a practical route toward scalable segmentation of multiphase pore-scale X-ray images.

The contribution of this study is therefore not only a new architecture, but also evidence that a single adapted model can support segmentation across multiple rock types, wettability states, and fluid systems within a common experimental framework. This does not remove the need for careful data curation, protocol definition, and validation on new domains, but it substantially reduces the dependence of pore-scale segmentation on manual correction and repeated model redevelopment and training. More broadly, the approach provides a basis for future work on broader lithology coverage, physics-aware segmentation, and time-resolved 4D pore-scale imaging.

\section*{Acknowledgments}
The first author gratefully acknowledges the support of the ‘111 Center’ (B25049) and the China Scholarship Council (CSC) for financial support.

The SAMamba3D code is available at  \url{https://github.com/ImperialCollegeLondon/SAMamba-3D}.

\bibliographystyle{unsrt}  
\bibliography{references}

\end{document}